\pdfoutput=1

\documentclass[11pt]{article}

\usepackage[final]{acl}
\usepackage{authblk}

\usepackage{times}
\usepackage{latexsym}
\usepackage{tcolorbox}
\usepackage{algorithm,algpseudocode}
\usepackage{amsmath}
\usepackage{subcaption}
\usepackage{fvextra} 

\usepackage{booktabs}
\usepackage{amssymb}
\usepackage{graphicx}
\usepackage[frozencache,cachedir=.]{minted}

\usepackage{titlesec}
\titlespacing{\paragraph}{0pt}{0.3\baselineskip}{0.1\baselineskip}

\usepackage{multirow}

\newenvironment{example}{%
    \vspace{0.5em}
    \begin{quotation}
}{%
    \end{quotation}
    \vspace{0.5em}
}

\newlength{\myMheight}
\newcommand{\hf}{\includegraphics[height=\myMheight]{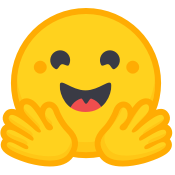}}

\usepackage{tcolorbox}
\tcbuselibrary{skins}
\newcommand{\improvedParbox}[1]{%
  \begin{tcolorbox}[
    enhanced,
    colback=white,
    colframe=gray!50!black,
    boxrule=0.5pt,
    arc=1mm,
    left=3mm,
    right=3mm,
    top=2mm,
    bottom=2mm,
    boxsep=0pt,
    width=\columnwidth
  ]
    #1
  \end{tcolorbox}
}

\usepackage[T1]{fontenc}
\usepackage[utf8]{inputenc}
\usepackage{microtype}

\title{
Logic Haystacks: Probing LLMs Long-Context Logical Reasoning\\
(Without Easily Identifiable Unrelated Padding)
}

\author[]{Damien Sileo}
\affil[]{Univ. Lille, Inria, CNRS, Centrale Lille, UMR 9189 - CRIStAL, F-59000 Lille, France}
\affil[]{\url{damien.sileo@inria.fr}}

\begin{document}
\maketitle
\begin{abstract}
Large language models demonstrate promising long context processing capabilities, with recent models touting context windows close to one million tokens. However, the evaluations supporting these claims often involve simple retrieval tasks or synthetic tasks padded with irrelevant text, which the models may easily detect and discard.
In this work, we generate lengthy simplified English text with first-order logic representations spanning up to 2048 clauses ($\approx25k$ GPT-4 tokens). We formulate an evaluation task with evidence retrieval for contradiction detection.  The long, homogeneous text is filled with distractors that are both hard to distinguish from relevant evidences and provably not interfering with them. Our evaluation of evidence retrieval shows that the effective context window is much smaller with realistic distractors, already crumbling at 128 clauses.

\end{abstract}

\section{Introduction}

Recent large language models (LLMs) can process long context windows, which expands the scope of their applications. However, the theoretical context length alone fails to capture a model's actual performance across varying input sizes \cite{liu2024lost}. Several benchmarks have been developed to evaluate the actual capabilities of these models in reasoning over extended contexts. 

Human annotated benchmarks   \cite{bowman2022quality,wang2024novelqa,bai2024longbench} are expensive to scale and constrained by both memory limitations and the finite attention span of annotators. On the other hand, synthetic datasets can be arbitrarily difficult, but they mainly use simple tasks (retrieval or reasoning) then drown the relevant inputs in unrelated text \cite{kamradt2023needlenih,levy2024same,li2024needlebench}. This stratagem is not only easy to counteract, as language models can detect the irrelevant input, but also risky, as some statements in BookCorpus or Paul Graham essays, which are routinely used, might interfere with the original problems.
Generating realistic distractors as padding is desirable, but it increases the risk of semantic collisions that would perturb the task.

In this work, we use grammars to generate simplified English paired with logical representations to create long input text while controlling the semantics of the text. We structure the generated expressions into $\textsc{Premise}_{[N]}$, \textsc{Hypothesis} pairs, where the premise is a conjunction of $N$ sentences referred to as clauses. We ask models to identify the premise clauses which contradict the hypothesis when taken together. We detect contradictions with a First-Order Logic (FOL) solver \cite{goodwin-etal-2020-probing}. A simple example is:
$\textsc{Premise}_{[3]}$: \textit{L0: Everyone who is happy is rich. L1: Mary is happy. L2: Nina is rich.} \textsc{Hypothesis}: \textit{Mary is not rich.}. \textsc{Contradiction Evidences}: L0, L1.

We use a previously proposed grammar for FOL with English correspondences to generate reasoning problems \cite{sileo2024scaling}, and we propose a methodology to scale premise generation to thousands of sentences. This is a challenging problem because naively scaling the problem causes paradoxes (e.g. \textit{Mary is happy. Paul is not rich. [...] Mary is not happy.}) at rapidly growing rates.
We also propose a method to isolate sufficient and necessary explanations that can be reliably used to probe LLM for evidence identification (detecting the causes of the contradiction).
Our contributions are as follows: (i) a scalable algorithm to generate formally verified reasoning datasets to probe complex logical reasoning inside long contexts. (ii) a method to extract necessary and sufficient evidence, providing a logic task formulation to probe LLMs' ability to explain a contradiction (iii) evaluation results for recent large language models, comparing padding with realistic distractors, and publicly available datasets\footnote{\href{https://hf.co/datasets/sileod/LogicHaystacks}{[data:HF-datasets \hf]}\label{footnote:urls}}.

\section{Scaling logical reasoning data generation or long contexts}

One of the main use cases of processing long inputs is piecing together relevant pieces of knowledge (e.g. during Retrieval Augmented Generation \cite{lewis2020retrieval}. Finding whether the evidence contradicts a fact (the hypothesis) is a realistic use case that we model with a large premise expressing logical statements with some linguistic variety.

We construct each premise as a conjunction of $N$ sentences paired with logical representation.
Then, for each premise, we search for single hypotheses that trigger contradictions\footnote{We focus on contradiction detection rather than entailment detection because we found that hypotheses not triggering contradiction were relatively rare, less than $30\%$ of cases.}. Our first goal is to generate long premises that are not paradoxical. There are many strategies to scale the number of clauses while avoiding contradictions, like increasing the number of predicate and variable names \cite{monasson1999determining}. However, we need the premise to contain many challenging distractors, so we still want to have symbols that occur multiple times to prevent clauses from being easily discarded.

\subsection{Satisfiable merging}

To maintain distractor difficulty with many premises, we propose a simple method to generate premises with many expressions that are satisfiable (non-contradictory) while containing relatively few concepts.
We leverage the Vampire \cite{reger2022vampire} theorem prover to check whether a formula is satisfiable (i.e. not contradictory). 

We start by generating $K$ formulas of 32 clauses (stage $i=0$).  We define a satisfiable merging as a conjunction of two formulas, where we check whether the conjunction is contradictory; if it is, we use the proof of the contradiction to identify the set of clauses triggering the contradiction. We sample one of these clauses, remove it, and we repeat the removals until the conjunction is satisfiable\footnote{MaxSAT algorithms could achieve this process more efficiently but are not available for full-fledged FOL. In addition, MaxSAT adds another constraint that we do not need, because we do not necessarily need maximality.}. We construct the $i+1$ stage by randomly selecting $K$ formulas pairs from stage $i$, and computing their satisfiable merge. Thus, the maximum formula size doubles at each stage. We continue until we get $K$ premises of up to 4096 clauses.

\subsection{Locating evidence with counterfactuals}

The Vampire theorem prover provides a derivation listing specific clauses used to support a contradiction.
However, this only provides a \textit{sufficient} set of evidence, not a \textit{necessary} set. There might be other clauses that also support the contradiction.
To identify necessary evidence, we look at the clauses in the premises that are supporting a proof.
For each evidence $e$, we recompute whether (\textsc{Premise} without $e$, \textsc{Hypothesis}) pair is contradictory. If the pair remains contradictory, this means that the evidence was not a necessary evidence of the contradiction. By doing that, we can keep only examples where all evidences are necessary.
For these examples, evidence retrieval has a unique solution and is the well-formed task that we use to probe logical reasoning in the presence of lengthy input.

\begin{figure*}[htb]
  \centering
  \begin{minipage}{0.5\textwidth}
    \centering
    \includegraphics[width=\linewidth]{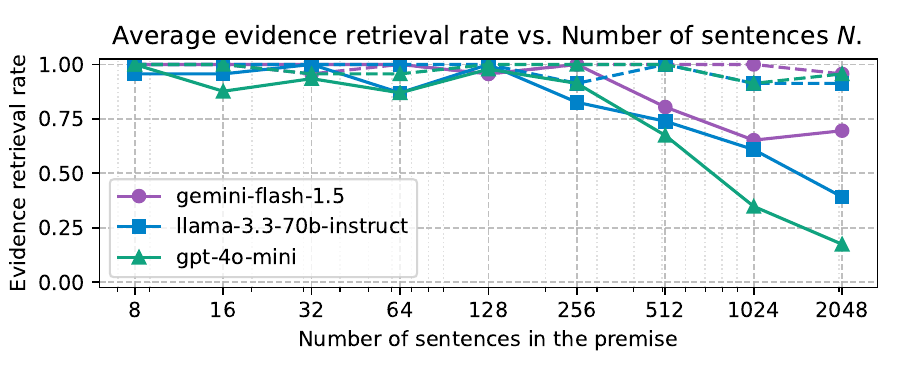}
    \subcaption{One evidence}\label{fig:zs1a}
  \end{minipage}%
  \begin{minipage}{0.5\textwidth}
    \centering
    \includegraphics[width=\linewidth]{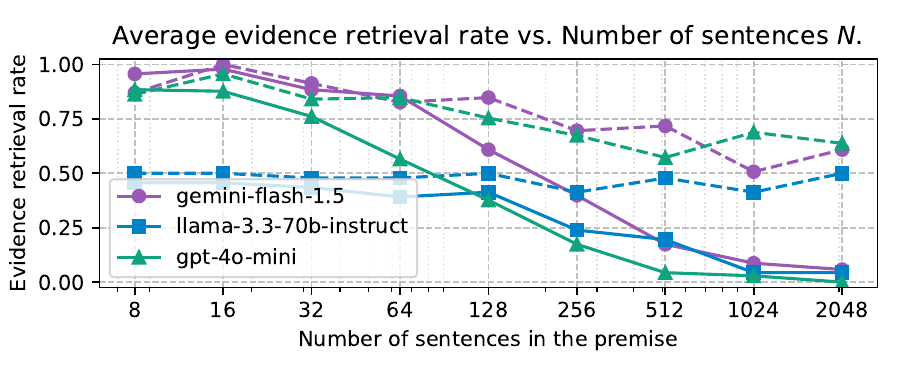}
    \subcaption{Two evidences}\label{fig:zs1b}
  \end{minipage}\par\medskip
  \begin{minipage}{0.5\textwidth}
    \centering
    \includegraphics[width=\linewidth]{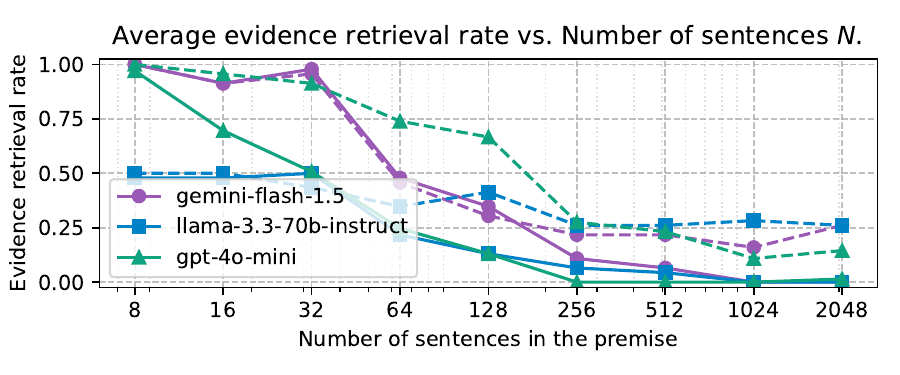}
    \subcaption{One evidence, not isolating the hypothesis}\label{fig:zs1c}
  \end{minipage}%
  \begin{minipage}{0.5\textwidth}
    \centering
    \includegraphics[width=\linewidth]{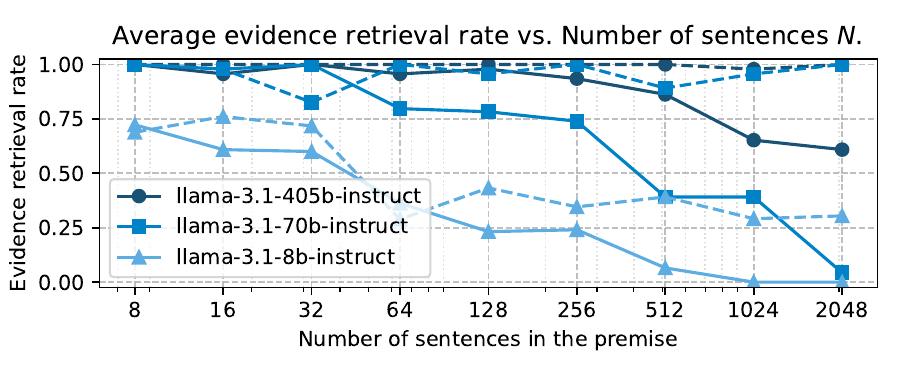}
    \subcaption{One evidence, impact of model size}\label{fig:zs1d}
  \end{minipage}
  \caption{Evidence retrieval on Logic Haystacks for 8 to 2048 clauses. For each model, the dashed line shows the accuracy when we replace the grammar-generated clauses with random sentences from Paul Graham essays.}
  \label{fig:zs1}
\end{figure*}

\subsection{Logic Haystacks dataset construction}

We leverage Unigram-FOL 
\cite{sileo2024scaling}, a parallel grammar-based generator of English expressions aligned to first-order logic expressions.
It generates facts with independent atomic propositions, or with independent predicates, and conditionals based on facts. Unigram-FOL grammar has higher transfer with the human-crafted FOLIO dataset \cite{han2022folio} and with human-crafted formal semantics constructions \cite{fragmentsaaai} when compared to other comparable grammars (RuleTaker \cite{ruletaker}, LogicNLI \cite{tian-etal-2021-diagnosing} and FLD \cite{fld23})

We use three times more names and predicates (78 instead of 26) to reduce the chance of semantic collisions at a manageable rate while having frequent symbol repetition.

We sample atomic predicates as hypotheses to avoid ambiguous constructs such as vacuous implications. We discard hypotheses that are in the premise with the same surface form to prevent superficial matching.

The generation process took 12 days on an Intel Xeon Gold 5320 CPU, parallelized over 52 threads.

We only consider the last stage with up to 4096 premises, and we discard all premises with less than 2048 premises.
Then, to obtain shorter hypotheses, we keep 8,16,...,2048 premises by randomly sampling the distractors. We could have used the earlier stages, but the proof structure statistics (depth, axioms used) depend on the stage and would prevent a controlled comparison.

We independently generate a total of 600 validation examples and 2790 test (starting with K=2000 and K=10000 respectively) examples including contradiction, entailment, and neutral labels, but we will only sample from the contradiction subset here, and consider examples with 1, 2, or 3 necessary and sufficient evidences. We sample 200 examples for each variation of the test set.

\improvedParbox{%
\textsc{Prompt Template}\label{Prompt A}: "Premise:
\{p\}
Hypothesis:\{h\}
Given the premise, find the \{k\} line identifiers explaining why the hypothesis is logically contradicted by the premise. Answer directly with no explanation using comma-separated line identifiers between <answer> and </answer> tags. Format illustration: "<answer>L42</answer>".).
}

We construct the \textsc{Prompt Template}, iterating on Llama-3.1-8B-Instruct with the 2-evidence validation data until we obtained a satisfactory output format.  We picked this model for prompt selection because it is the weakest model we used, and all other models conformed to the output format upon inspection and automated verifications.

Chain of thoughts did not drastically change the results in our early experiments.  We parse the output automatically to compute the evidence retrieval rate for each example. We evaluate the retrieval accuracy with the Jaccard similarity metric between predicted evidence and ground truth.

We evaluate off-the-shelf instruction-tuned language models API. We evaluate the Llama 3 family (FP8), Gemini 1.5-002 Flash, and GPT-4o-mini, with the default hyperparameters. These models all achieve near-perfect Needle-In-Haystack accuracy with at least 32k tokens.

\subsection{Results}

Figure \ref{fig:zs1} presents retrieval scores across different configurations. We use dotted lines when reporting results using Paul Graham sentences \footnote{\href{https://hf.co/datasets/sgoel9/paul_graham_essays}{hf.co/datasets/sgoel9/paul\_graham\_essays}} as distractors, following prior work \cite{kamradt2023needlenih,li2024needlebench,levy2024same}. We split the essays into sentences with the pysbd \cite{sadvilkar-neumann-2020-pysbd} tokenizer.

\paragraph{One evidence (\ref{fig:zs1a})}
 Even when the proof only involves one evidence, we see a high difference between the type of distractors. Gemini Flash performs much better than the compared models with the realistic distractors, but that wouldn't have been conspicuous with easier padding.
\paragraph{Two evidences (\ref{fig:zs1b})} With two evidences, the performance drops significantly for all models, as none of them obtain a retrieval rate above $20\%$ with 1024 clauses.
\paragraph{Not isolating the hypothesis (\ref{fig:zs1c})}
We also insert the hypothesis at a random position inside the premise and remove it from the prompt\footnote{We rephrase the prompt with \textit{Find $\{k\}$ line identifiers that logically contradict each other within the premise.}}. This makes it much harder to notice logical contradictions. Interestingly, this task is still hard even with the easily identifiable distractors.


\paragraph{Model size (\ref{fig:zs1d})}
We evaluate the Llama 3.1 model family (8B, 70B, 405B) to evaluate the effect of model size in a controlled experiment. While model size helps, it is not sufficient to work even with 1 distractor, revealing potentially fundamental issues in this generation of models.

\subsection{Qualitative error analysis}
We show randomly sampled error examples in the Appendix \ref{appendixB}. We notice that the incorrectly predicted evidences are lexically related without constituting clear-cut logical contradictions. It is also interesting to note that when using Paul Graham essays as padding, the models often select them, but that they do not constitute clear-cut contradictions.

\section{Related work}

Our work stands at the intersection of two lines of research: logical reasoning with synthetic simplified English data, and context length evaluation. \citet{levy2024same} and \citet{kuratov2024babilong} explored the intersection of the two, but they imported the padding technique from previous work on LLM stress-texting, while we scale the dataset's original generation process instead of padding it with other text.
Human annotated long context benchmarks are very valuable but hard to annotate \cite{bowman2022quality,wang2024novelqa,bai2024longbench}, which causes current language models to saturate them.

\paragraph{Synthetic datasets for reasoning}


Numerous works investigate the logical capabilities of NLP models using textual datasets and symbolic reasoning \cite{helwe2022logitorch}. We focus on the grammar-derived synthetic datasets. RuleTaker \citep{ruletaker} LogicNLI \citep{tian-etal-2021-diagnosing}, FLD \cite{fld23} and FOL-NLI \cite{sileo2024scaling} address different subsets of first-order logic with English translations. Other works also explore non-standard logic with synthetic datasets, notably probabilistic \citep{jin2023cladder,sileo2022probing}, paraconsistent \cite{kazemi2024boardgameqa}, epistemic \cite{sileo-lernould-2023-mindgames} logics.

These approaches focus on input sizes typically suitable to a standard BERT \cite{devlin2018bert} encoder (<512 tokens). Here, we push the number of expressions in the input while avoiding paradoxes. This is related to the satisfiability problem which was explored by \citet{fragmentsaaai,richardson2022pushing} who use a solver to study the satisfiability in natural language using the Z3 solver and dedicated generation logic on constrained problems.  However, they also focus on relatively moderate text size while we use satisfiability checking as a stepping stone to generate large text and not only as a task in itself.

\paragraph{LLM context length stress tests}  Our work is also related to context window stress testing. The Long-Range Arena \cite{tay2021long} provides the first systematic analysis of the long-range processing capabilities of text encoders, focusing mainly on algorithmic reasoning and retrieval tasks. Needle in Haystack benchmarks \cite{kamradt2023needlenih,li2024needlebench} test longer window sizes with simple retrieval tasks and use Paul Graham essays as padding. BABILong   \cite{kuratov2024babilong} uses bAbi \cite{babi} reasoning tasks and interleaves relevant text with irrelevant input from BookCorpus \cite{Zhu_2015_ICCV}. FlenQA \cite{levy2024same} applies a similar process to the RuleTaker \cite{ruletaker} deductive logical reasoning task and uses Paul Graham essays as padding. Ruler uses simple algorithmic tasks like variable tracking and word counting. They also use Paul Graham essays as noise, or repetitions of sentences such as \textit{The sky is blue}, following \citet{mohtashami2024random}. 
The MuSR dataset uses GPT-4 generated \cite{sprague2023musr} problems, which makes it hard to verify the problems' integrity at scale.

\section{Conclusion}
Evaluation datasets are critical to guide language model construction. We proposed methods to scale logic datasets for long context probing. We confirm that using external text as padding leads to over-estimating the context window and propose a new dataset, Logic Haystacks, that can provide an unbiased signal to evaluate long-context processing architectures. Further work is needed to scale the generation process to training data and to evaluate whether training on long-context logical reasoning synthetic data (using a less expressive logic to scale data construction more easily, because going beyond 4096 clauses starts being intractable with our method) leads to transferable gains on Logic Haystacks.

\section*{Limitations}
While our work provides valuable insights into long-context logical reasoning, several limitations should be noted. The generation process is computationally expensive, taking 12 days even with parallelization, which limits the scale of datasets that can be produced. While our method uses simplified English and a constrained subset of first-order logic to enable formal verification, this may not fully capture the complexity and nuance of natural language reasoning. The focus on contradiction detection rather than other logical relations like entailment or equivalence narrows the scope of evaluation. Additionally, our zero-shot evaluation on a limited set of models with default hyperparameters may not fully represent the potential of these systems, particularly under different prompting strategies or with tuned parameters. Finally, while the lack of human performance benchmarks could be seen as a limitation, it's worth noting that our task deliberately pushes beyond typical human cognitive limits, as processing thousands of interrelated logical statements is precisely the kind of task where we expect machines to surpass human capabilities.

\ifdefined\acl@finalcopy
\section*{Acknowledgement}
This work was supported by the French National Research Agency (ANR) through the ANR-24-CE23-4637 grant (Adada project).

\fi 
\bibliography{anthology,custom}

\onecolumn 
\appendix
\renewcommand\thesection{\Alph{section}} 
\section{Appendix A - Prompt example\label{appendixA}}
\begin{Verbatim}[breaklines=true, breaksymbolleft=, breaksymbolright=]

Premise:
L0: if someone is richer than Jennifer then he/she is a happy person and vice versa
L1: Susan and Delbert hate each other
L2: more than one person in the room is a romantic person
L3: ““Mirrors do not fog permanently in one particular house.” or “A tower does not lean significantly without falling.” or both” if “Dana is quieter than Beatrice” and vice versa
L4: not everyone in the room who frequently participates in hackathons and coding competitions does not own a smart tv
L5: Walter is not a brave person
L6: Leonard is younger than Justin
L7: Dcalvina and Susan hate each other
L8: someone in the room does not hate Benjamin
L9: Jason is a client of John
L10: Stephen is allergic to anything
L11: Dorothy is a client of Elsie
L12: Janna neither is a client of Ramon nor is older than Ossie
L13: Dorothy is not an avid collector of autographed memorabilia from famous musicians
L14: everyone in the room who does not enjoy mountain biking hosts a popular podcast about emerging technologies
L15: not everyone in the room does not collect vintage vinyl records or is a curious blue eyed tourist
L16: Dorothy, Brian, Natividad, Natasha, Jason, Michael, Napoleon, Dcalvina, Melissa, Justin, Chae, Charlette, Rex, Pamela, Janna, Elsie, Calvin, Elizabeth, Bernard, Michelle, Delbert, Dana, Caitlin, Alice, Beatrice, Brandon, James, Joseph, Francis, Brenda, Chae, Virginia, Dionne, Louise, Christopher, Nelson, Walter, Ramon, Carol, Philomena, Raymonde, Stephen, Carla, Shirley are the only persons in the room.
L17: everyone in the room who is not a night owl is not liked by Gary
L18: everyone outside the room is a tall curly_haired strong tourist if they is a cybersecurity expert
L19: at least two persons in the room is a colorblind european
L20: someone in the room does not hate Jennifer
L21: Rex and John hate each other
L22: Shirley is younger than Steven
L23: Justin is quieter than Genevieve
L24: “A tower leans significantly but never falls.” or “Glass rain falls on a distant planet.” but not both
L25: everyone in the room is younger than Jeannine if they is not liked by Benjamin and vice versa
L26: it is not the case that “Michelle is richer than Shirley”
L27: at least one person in the room is liked by Lisa
L28: everyone outside the room enjoys rooftop gardening if they owns a high-end gaming PC with custom-built components
L29: everyone in the room who is younger than Genevieve owns a 3D printer
L30: if “it is not the case that “Shirley is a sibling of Genevieve”” then ““Mirrors fog permanently in one particular house.” and “A city has not outlawed the use of round tables.””
L31: Charlette and John like each other.
Hypothesis:
Dorothy is an avid collector of autographed memorabilia from famous musicians

Given the premise, find the 1 evidence explaining why the hypothesis is contradicted by the premise. Answer directly with no explanation and only with comma-separated line ids, e.g., "L0,L3."

\end{Verbatim}
GPT-4o-mini prediction: "L12".\\
Correct answer: "L13."\\
We note that at this scale, most models solve the task very well.

\newpage
\section{Appendix B - Error examples\label{appendixB}}

Error examples using Gemini 1.5 Flash with 2048 clauses:

\begin{example}
\textbf{Hypothesis:}\\ David and Christopher hate each other

\textbf{Predicted Evidence:}\\
L266: Kathleen is a sibling of Tamara

\textbf{Ground truth:}\\
L126: Christopher and David like each other
\end{example}

\begin{example}
\textbf{Hypothesis:}\\ Marvin is a formal european

\textbf{Predicted Evidence:}\\
L26: Michelle and Marvin hate each other

\textbf{Ground truth:}\\
L587: Marvin is not curious, not formal
\end{example}

\begin{example}
\textbf{Hypothesis:}\\ Jewell travels domestically frequently

\textbf{Predicted Evidence (Paul Graham Distractors):}\\
L299: Once you start considering this question, you have opened a real can of worms.

\textbf{Ground truth:}\\
L729: Jewell does not travel domestically frequently
\end{example}

\begin{example}
\textbf{Hypothesis:}\\ Donald is a sibling of Michael

\textbf{Predicted Evidence (Paul Graham Distractors):}\\
L80: Michael does enjoy trail running\\
L87: Certainly schools should teach students how to write.

\textbf{Ground truth:}\\
L80: Michael does enjoy trail running\\
L987: Michael either is a sibling of Donald or does enjoy trail running but not both
\end{example}

\end{document}